# 3D Facial Geometry Recovery from a Depth View with Attention Guided Generative Adversarial Network

Xiaoxu Cai, Hui Yu, *Senior Member, IEEE*, Jianwen Lou, Xuguang Zhang, Gongfa Li, Junyu Dong, *Member, IEEE*

*Abstract*—We present to recover the complete 3D facial geometry from a single depth view by proposing an Attention Guided Generative Adversarial Networks (AGGAN). In contrast to existing work which normally requires two or more depth views to recover a full 3D facial geometry, the proposed AGGAN is able to generate a dense 3D voxel grid of the face from a single unconstrained depth view. Specifically, AGGAN encodes the 3D facial geometry within a voxel space and utilizes an attention-guided GAN to model the ill-posed 2.5D depth-3D mapping. Multiple loss functions, which enforce the 3D facial geometry consistency, together with a prior distribution of facial surface points in voxel space are incorporated to guide the training process. Both qualitative and quantitative comparisons show that AGGAN recovers a more complete and smoother 3D facial shape, with the capability to handle a much wider range of view angles and resist to noise in the depth view than conventional methods.

*Index Terms*—3D facial geometry recovery, depth view, GAN

## I. INTRODUCTION

A NUMBER of artificial intelligent systems such as robots and agents are designed for interacting with humans via multiple facial sensing techniques and learning methods. In some of those systems, reconstructing 3D facial geometry from integrated depth sensors is a fundamental step to achieve accurate facial expression capture and recognition [1], [2]. With the continuously increasing sensing precision and portability, depth camera is becoming a critical tool in capturing 3D objects including the human face. For example, the Apple's TrueDepth camera has been successfully deployed in mobile devices (Iphone X) to support 3D facial applications. This motivates an important research stream which aims to reconstruct 3D facial geometry from 2.5D depth views. Existing methods [3]-[5] were able to obtain the promising 3D shape by fusing multiple views of depth maps. However, it is not applicable for the practical application because of the complexity of multiple depth maps acquisition. Compared with these approaches, recovering geometry from a single view is

Xiaoxu Cai, Hui Yu and Jianwen Lou are with the School of Creative Technologies, University of Portsmouth, Portsmouth, PO1 2DJ, UK.

Xuguang Zhang is with the School of Communication Engineering, Hangzhou Dianzi University, Hangzhou, 310018, China.

Gongfa Li is with the Key Laboratory of Metallurgical Equipment and Control Technology, Ministry of Education, Wuhan University of Science and Technology, Wuhan 430081, China.

Junyu Dong is with the College of Information Science and Engineering, Ocean University of China, Qingdao, 266100, China.

Corresponding Author: Hui Yu, Email: hui.yu@port.ac.uk.

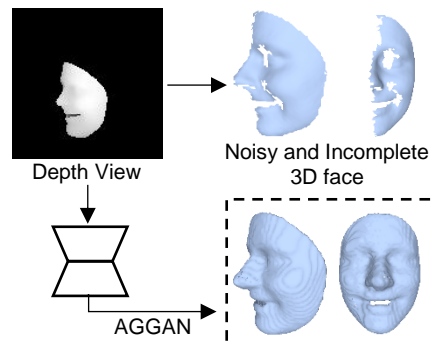

Fig. 1. AGGAN can recover the complete 3D facial geometry from a noisy and non-frontal depth view.

more feasible and convenient in real applications. Nevertheless, it is very challenging to recover 3D facial geometry precisely if there is only one depth view available. This is mainly because partial observation can be theoretically associated with an infinite number of possible 3D facial information, especially when the depth view is non-frontal with the depth information of the occluded facial parts missing (see Fig. 1). The problem above can be interpreted as reconstructing a facial surface from 3D point cloud projected from the given depth view. This is a long-lasting research topic that has been extensively studied in computer graphics [6]-[12]. Typical solutions reconstruct the surface by either fitting the points with a discrete grid [7], [8] or using the zero set of an implicit function [9]-[11] such as the indicator function defining the interior and exterior of the object surface. However, these approaches degenerate sharply when dealing with noisy and non-frontal depth views, and normally can only recover partial 3D facial geometry. The problem can also be cast to 3D shape non-rigid registration [13]-[18] which is also pervasive in computer graphics. Generally, non-rigid registration methods first build dense point correspondences between the projected 3D point cloud and a template 3D facial mesh, and then conforms the template mesh to the point cloud using the built correspondences. Whereas a complete 3D facial geometry can be acquired with such methods, facial parts occluded or missing in the depth view can rarely be warped correctly on the template because false correspondences are prone to being found for them. Furthermore, these methods usually require certain hand-selected facial feature points to rigidly align the template with the point cloud for a promising registration initialization. In summary, existing methods can hardly handle imperfections in the given depth view such as the noise and missing data.

We have found that existing methods merely utilize noisy 3D

information embedded in the given imperfect depth view, while making no attempt to build and exploit a 3D facial point distribution which covers various facial geometries. With such a distribution, the reconstruction problem becomes generating or sampling 3D facial points from that distribution given a depth view as a conditional input, which could be solved efficiently by Generative Adversarial Networks (GAN) [19]. Accordingly, in this study we propose AGGAN which is a variant of GAN to learn the highly-complicated conditional distribution of 3D facial geometry given its depth view from thousands of synthetic depth-3D pairs. First, we encode the 3D facial geometry within a high-resolution voxel grid which has shown robustness in depicting 3D shapes [20]-[23]. We then guide the GAN to extract features that are more sensitive and discriminative in locating 3D facial points by incorporating the attention mechanism which has been validated in many other computer vision tasks [24]-[26]. To build a generative model covering a variety of natural depth-3D mappings, large variations in head pose and facial expression together with random noise are introduced during synthesizing training depth views.

Compared with existing methods on data generated from benchmark facial image datasets, the proposed data-driven AGGAN recovers a more complete and smoother 3D facial shape, while being able to handle a much wider range of view angles and more resistant to noise in the input depth view. Overall, our main contributions are as follows:

- To the best of our knowledge, this is the first work of its kind that utilizes GAN to recover 3D facial geometry from a single unconstrained depth view.
- We demonstrate that the incorporation of the attention mechanism into GAN can improve the precision of 3D facial geometry prediction.
- We showcase that using synthetic facial depth views for training is helpful in generalizing AGGAN to real depth views captured from depth cameras.

## II. RELATED WORK

### A. 3D Surface Reconstruction from Point Cloud

The area of 3D surface reconstruction has witnessed impressive progress in the last two decades [12]. From the perspective of the reconstruction output, the proposed solutions can be broadly divided into two categories, producing either a discrete surface [7], [8] or an implicit function [9]-[11]. The first kind of solutions typically fits a regular grid to the given points such as the well-known Marching Cubes [7], [8] which extracts the surface by finding intersections between the cubes of the grid and the points. The latter type utilizes the knowledge of the exterior and interior of the surface with an implicit function for reconstruction. The implicit function can have various forms such as a signed distance field [9] or an indicator function [10], whereby the reconstructed surface is found by isocontouring for an appropriate isovalue. However, when the point density is low, there are outliers or missing data, these methods are prone to generating an incomplete surface that poorly approximates the desired object shape. As a result, they can hardly deal with a single unconstrained facial depth view which is often noisy and with a head pose.

### B. 3D Shape Non-rigid Registration

The concerned reconstruction problem can be projected into the 3D non-rigid registration framework [13]-[16] if there is a facial geometry prior available. A typical solution is to register a facial template mesh to the given depth view using a deformation model based on smooth local affine transforms. Primarily, the registration process has to estimate reliable correspondences between the template and 3D points projected from the depth view for warping the template to match the underlying geometry of the captured depth data. False correspondence can cause strong shape distortions that are inconsistent with the desired facial shape. However, such correspondences are inaccessible when the given depth view is noisy and non-frontal with partial facial regions occluded. Moreover, a promising correspondence estimation often requires hand-selected facial feature correspondences [13], [14] or a rigidly-aligned shape prior [15]-[17], that offers a strong approximation of the target facial geometry. This is against with the most general setting where no facial geometry prior and feature point correspondences are available. All these issues make 3D reconstruction from a single unconstrained depth view intractable with existing non-rigid registration methods.

### C. 3D Reconstruction from a Single Depth View with Deep Learning

Whereas learning the 3D facial shape from a single depth view with data-driven deep neural networks remains almost unexplored, there are several studies [21], [27]-[31] working on single depth view 3D object reconstruction. However, the early approaches[27], [28] apply a low resolution voxel grid ($\leq 40 \times 40 \times 40$) which can only preserve the coarse shape information of the object. To solve this problem, Dai et al. [29] propose a two-stage pipeline: first using the neural network to predict a shape prior encoded with a $32 \times 32 \times 32$ voxel grid from the given depth view, then synthesizing a higher resolution shape based on a pre-built shape database. Such a shape database is however very difficult to construct, especially for the human face which has extensive shape variations. The SSCNet [30] extends the reconstruction to 3D indoor scene which contains multiple object categories and requires a much higher-resolution volumetric space for representation. The method leverages the synthetic scene data which provides both the depth view and the ground-truth voxel-level occupancy annotations, which significantly reduces the expense for collecting the high-resolution training data. Inspired by these studies, we propose to solve the ill-posed single depth view 3D face reconstruction with deep neural networks. To model the complex non-rigid facial shape motions and deformations within the network, we synthesize a large amount of training data by altering along the dimensions such as facial identity, expression and head pose.

## III. METHODOLOGY

In contrast with existing methods focusing on modelling only the given imperfect depth data, we propose to solve the ill-posed single depth view 3D face reconstruction in a more data-driven manner. Specifically, we propose an attention-guided GAN named as AGGAN (see Fig. 2) to model the complex 2.5D depth-3D relationship by learning from a large amount of synthesized training pairs. The generator of AGGAN approximates the real conditional distribution of 3D facial surface given its depth view. This data-driven prior is supposed to be more robust than manually specialized priors (e.g. distance field function [9], indicator function [10] or template 3D facial mesh [13], [14]) used in previous methods on addressing challenging data imperfections such as noise, missing/occluded facial parts. In the following sessions, we will discuss in detail the proposed AGGAN and the training data synthesis.

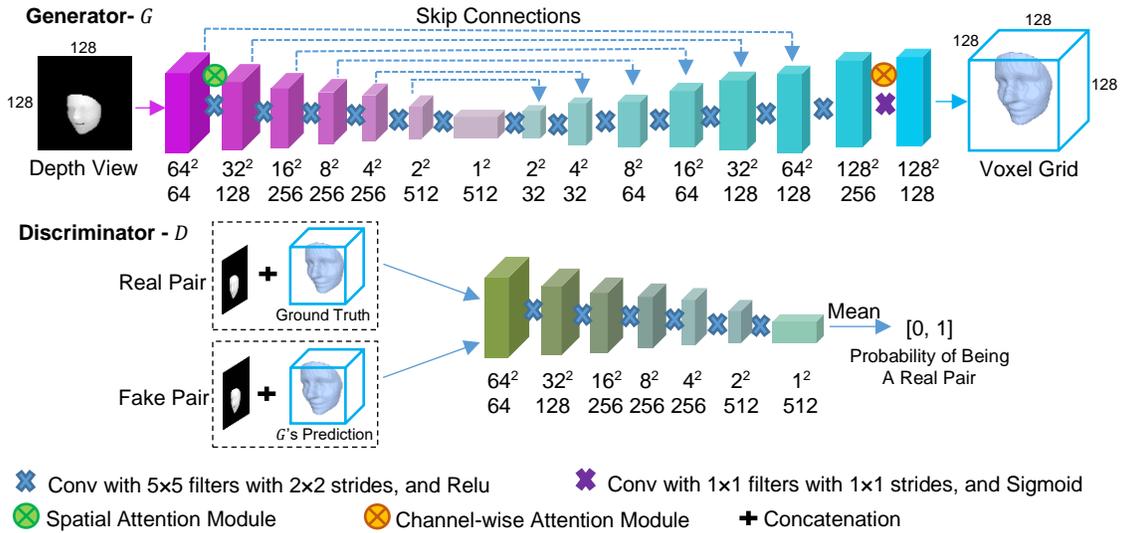

Fig. 2. The architecture of AGGAN.

### A. AGGAN

From previous work [20], [21] on 3D shape reconstruction, the voxel representation shows a promising ability in depicting 3D geometry and can be seamlessly processed by deep neural networks. We thus encode the 3D facial geometry within a 3D voxel grid whose voxel occupancy (1 for facial point and 0 for non-facial point) indicates if the current point belongs to the facial surface or not. The voxel grid resolution is set as $128 \times 128 \times 128$ which was determined after balancing the grid's representation capability and the network's processing consumption.

Fig. 2 illustrates the structure of AGGAN. During training, the generator $G$ tries to learn the ground-truth 3D voxel grid which encodes the facial geometry from a $128 \times 128$ facial depth view. Coupling with the corresponding depth view, both $G$'s prediction and its ground truth counterpart are then fed into the discriminator $D$ for training a classifier to distinguish real reconstruction pairs (the pair of a depth view and its ground-truth voxel grid) from fake reconstruction pairs (the pair of a depth view and its $G$ prediction). $G$'s outputs are forced to not only get close to the ground truth voxel grid but also maximize the probability of $D$ making a mistake. This adversarial learning drives $G$ to recover a faithful 3D facial geometry that matches the input depth view. Given a new facial depth view, $G$ will be called to predict the 3D voxel grid that encodes the facial geometry.

#### 1) Generator and Discriminator

The generator is a fully convolutional encoder-decoder network with skip-connections. The encoder consists of seven convolutional layers, each of which uses a bank of $5 \times 5$ filters with $2 \times 2$ strides and is followed with a Leaky ReLU activation. Without specification, the remaining network applies the same filter setup. From the first convolutional layer to the last one, the number of feature map channel is 64, 128, 256, 256, 256, 512 and 512 respectively. On the other side, the decoder comprises eight transpose-convolutional layers, the first seven of which are followed with Leaky ReLU activations, while the last one is followed with a sigmoid function to regulate the final output as the voxel occupancy probability. The number of each transpose-convolutional layer's output channel is 32, 32, 64, 64, 128, 128, 256 and 128. The last transpose-convolutional layer is for fine-tuning purpose and uses a bank of $1 \times 1$ filters with $1 \times 1$ strides. Skip-connections are built between encoder and decoder to guarantee the information sharing and prevent the gradient vanishing problem.

The discriminator accepts a $128 \times 128 \times 129$ tensor concatenated by a facial depth view and a 3D voxel grid as input, and outputs a single scalar whose value is between 0 and 1 to specify the probability that the voxel grid fully matches the depth view. Excluding the input and the last layer, it has a same structure as the generator's encoder. The last layer calculates the mean of a $1 \times 1 \times 512$ feature vector output from the previous layer. This mean feature is shown effective in stabilizing the adversarial training [21].

#### 2) Attention Mechanism

**Spatial Attention.** In general, the face occupies only a partial region in the depth view. The left background region is noisy and might mislead the neural network to learn less informative features for 3D facial geometry prediction. To force the net-work to focus more on the foreground facial region during feature learning, we incorporate a spatial attention mechanism [22] into AGGAN's generator. After the first activation layer of the generator's encoder, two convolutional layers followed with a softmax function are applied on the low-level feature maps to generate a spatial weighting map (see Fig. 3):

$$SA = F_{sa}(f^l, W_{sa}) \tag{1}$$

$$F_{sa}(f^l, W_{sa}) = softmax(cv2_{sa}(cv1_{sa}(f^l, W_{sa}^1), W_{sa}^2)) \tag{2}$$

where $f^l \in \mathbb{R}^{C \times HW}$ stacks $C$ reshaped $1 \times HW$ low-level feature vectors output from the previous layer, $F_{sa}$ is the mapping function whose parameters are denoted as $W_{sa}$ and $SA$. $SA$ refers to the generated $H \times W \times 1$ spatial weighting map. $cv1_{sa}(\cdot)$ and $cv2_{sa}(\cdot)$ represent two convolutional layers which use $\frac{C}{8} C \times 1$ filters and a $\frac{C}{8} \times 1$ filter respectively, and whose parameters are $W_{sa}^1$ and $W_{sa}^2$. $softmax(\cdot)$ refers to the Softmax function. The final outputs of the spatial attention module can be obtained by weighting each previous feature map with $SA$ (see Fig. 3).

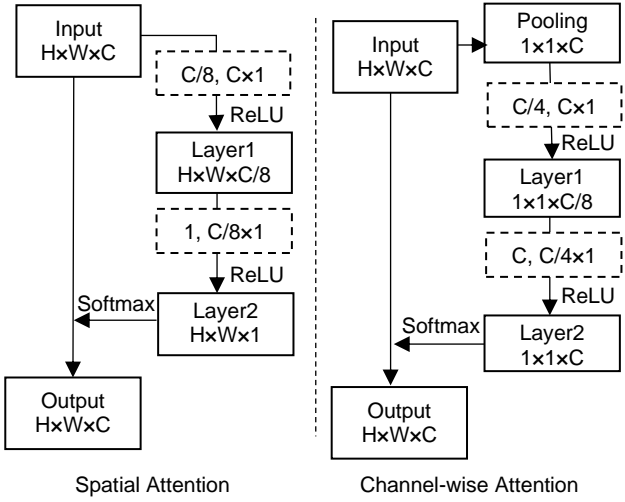

Fig. 3. The attention modules in AGGAN.

**Channel-wise Attention.** As reported in previous studies [22], different feature channels generated within convolutional neural networks correspond to different semantic information. We hence propose to incorporate the channel-wise attention mechanism into AGGAN to weight heavier on feature channels that show higher relevance in predicting 3D facial voxel grid. The channel-wise attention module is adhered to the second-to-last transpose-convolutional layer of the generator's decoder, aiming to produce a weighting vector for feature channels (see Fig. 3):

$$CA = F_{ca}(f^p, W_{ca}) \tag{3}$$

$$F_{ca}(f^p, W_{ca}) = softmax(cv2_{ca}(cv1_{ca}(f^p, W_{ca}^1), W_{ca}^2)) \tag{4}$$

where $f^p \in \mathbb{R}^{C \times 1}$ is the feature vector obtained by max-pooling feature maps output from the previous layer, $F_{ca}$ is the mapping function whose parameters are denoted as $W_{ca}$ and $CA$ is the generated $1 \times 1 \times C$ channel weighting vector. $cv1_{ca}(\cdot)$ and $cv2_{ca}(\cdot)$ represent two convolutional layers which use $\frac{C}{4} C \times 1$ filters and $C \frac{C}{4} \times 1$ filters respectively, and whose parameters are $W_{ca}^1$ and $W_{ca}^2$. $softmax(\cdot)$ refers to the Softmax function. Then, each previous feature map is weighted by the specific channel weighting value in $CA$ (see Fig. 3).

### 3) Objective Functions

The overall objective function of AGGAN consists of two parts: an adversarial loss $\mathcal{L}_{adv}$ for the whole network and an additional 3D face reconstruction loss $\mathcal{L}_{recons3d}$ for the generator.

**Adversarial Loss - $\mathcal{L}_{adv}$.** To train a generator that is able to predict an accurate 3D voxel grid $y$ from a depth view $x$, we apply a loss function as shown in (5). For the discriminator, the well-known WGAN-GP [32] loss function is adopted (see (6)):

$$\mathcal{L}_{adv}^g = -\mathbf{E}[D(y|x)] \tag{5}$$

$$\mathcal{L}_{adv}^d = \mathbf{E}[D(y|x)] - \mathbf{E}[D(\hat{y}|x)]$$
$$+ \lambda \mathbf{E}\left[\left(\left\|\nabla_{y'} D(y'|x)\right\|_2 - 1\right)^2\right] \tag{6}$$

where $\hat{y}$ is the ground-truth 3D voxel grid corresponding with the input depth view $x$ and $y' = \epsilon \hat{y} + (1 - \epsilon)y$, $\epsilon \sim U[0,1]$. $\lambda$ balances between optimizing the gradient penalty and the original objective in WGAN.

**3D Face Reconstruction Loss - $\mathcal{L}_{recons3d}$.** Since the face only occupies a small part of the overall volume, most voxels in the grid tend to be empty and the estimated voxel occupancy is prone to false positive. Inspired by this observation, we utilize a modified binary cross-entropy loss function [21], [33] to weight the penalty on false positive estimations and the penalty on false negative estimations in terms of the ratio of occupied voxels in the ground truth grid:

$$\mathcal{L}_{recons3d}^{ce} = -\sum_{i=1}^{h \times w \times d} \begin{bmatrix} (1 - \omega)\hat{y}_i \log y_i + \\ \omega(1 - \hat{y}_i) \log(1 - y_i) \end{bmatrix} \tag{7}$$

where $h$, $w$, $d$ is the voxel grid's height, width and depth respectively. For voxel $i$, $\hat{y}_i$ is the ground truth occupancy state and $y_i$ is the estimated occupancy state. $\omega$ denotes the ratio of occupied voxels in the ground truth grid. To further avoid false positive estimations, we impose a $L1$ sparsity constraint on the predicted voxel grid $y$:

$$\mathcal{L}_{recons3d}^{sparse} = |y|_1 \tag{8}$$

Overall, the loss functions for generator and discriminator in AGGAN are as follows:

$$\mathcal{L}_G = \alpha \mathcal{L}_{adv}^g + \beta \mathcal{L}_{recons3d}^{ce} + \gamma \mathcal{L}_{recons3d}^{sparse} \tag{9}$$

$$\mathcal{L}_D = \mathcal{L}_{adv}^d \tag{10}$$

where $\alpha$, $\beta$ and $\gamma$ are used to balance different loss terms, and their values are set empirically.

### B. Data Synthesis

Collecting real facial depth views and their precise 3D data in a volume sufficient for training a deep network is laborious and expensive. However, it's easy to get a depth view given a 3D face, head pose and camera projection matrix. Considering there are many high-quality 3D face datasets [34], [35] which cover a wide range of facial identities and expressions, we propose to synthesize depth views from the known 3D facial data for training and validating AGGAN.

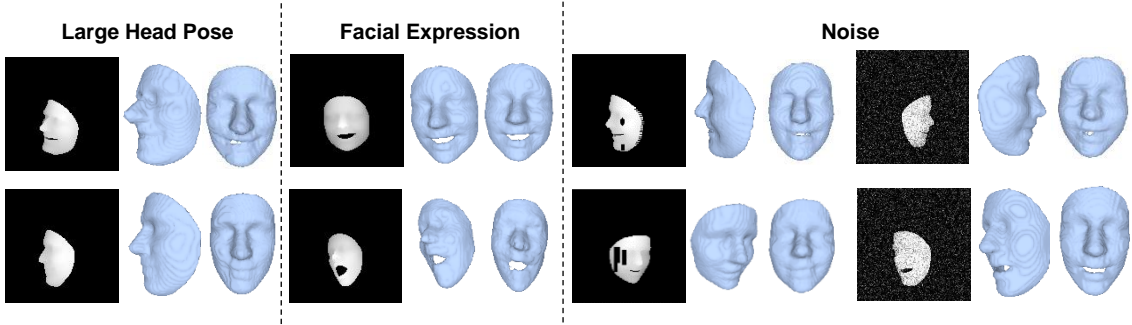

Fig. 4. Example results of AGGAN for depth views with large head pose, facial expression and noise.

TABLE I
IoU AND CE VALUES OF TESTING RESULTS

|  | IoU | CE |
|---|---|---|
| **AFW (10414 samples)** | 0.9916 | 0.0517 |
| **IBUG (3572 samples)** | 0.9937 | 0.0490 |
| **LFPW (33112 samples)** | 0.9913 | 0.0523 |

We use the dataset - 300W-LP proposed in [35] for data synthesis. 300W-LP contains in-the-wild face images from four independent benchmark databases including HELEN [36], LFPW [37], IBUG [38], AFW [39], and their 3D faces reconstructed by 3DMM fitting [35]. The reconstructed 3D faces capture the facial identity and expression exhibited in the images well and are represented with triangulated meshes that have a uniform topology. To introduce more variations in head pose, 300W-LP rotated the reconstructed 3D faces with multiple view angles and generated the corresponding RGB face images through image warping. This yields a dataset which contains more than 122K face images and their corresponding 3D face data. 300W-LP also provides the weak perspective projection to align each 3D facial mesh with the face in the image:

$$V_p = f \times P_r \times R \times V + T_{2d} \qquad (11)$$

where $V_p$ is the projected 3D face with its depth channel removed, $V$ is the reconstructed 3D face, $R$ is the rotation matrix, $P_r$ is the orthographic matrix $\begin{pmatrix} 1 & 0 & 0 \\ 0 & 1 & 0 \end{pmatrix}$, $f$ is the scale factor and $T_{2d}$ is the translation vector defined on the 2D image plane.

With (11), the 2D image pixel coordinates of each 3D facial vertex can be easily found. For a pixel in the depth view, we first find out the 3D vertex that is projected onto it and visible to the camera using Z-Buffer, then fill in the pixel value with the found 3D vertex's depth value. A depth view aligned with the face image can be acquired after going through all pixels with the operation above. To reduce the size of AGGAN for more efficient training, all synthetic depth views are resized to $128 \times 128$ and the aligned 3D facial meshes are shrunk accordingly. Considering real-world depth views are noisy, random Gaussian noise is further added to the synthesized depth views. For the facial mesh, we remove the neck and the ear part to focus on the main face region. The resulting mesh contains about 35K vertices. Inspired by previous work on 3D shape reconstruction [20], [21], we use a voxel grid to preserve the 3D facial geometry. In particular, the facial mesh is voxelized to a $128 \times 128 \times 128$ grid aligned with the depth view. Comparing with the vector, the voxel grid models 3D geometry in a way much closer to the real-world representation.

## IV. EXPERIMENTS

### A. Experimental Setup

**Datasets.** Depth views synthesized from HELEN are used for training AGGAN, while the rest depth views synthesized from 300W-LP [35] are used for testing. In total, there are 75,352 training samples and 47,098 testing samples. As mentioned in *Data Synthesis*, 300W-LP includes a variety of natural facial expressions and has been augmented to cover a wide range of head poses, e.g. with yaw angles ranging from $-90°$ to $90°$. The synthetic depth views have also been perturbed with random Gaussian noise to further simulate imperfections in real depth views.

**Implementation Details.** The generator and discriminator of AGGAN are optimized in an alternate manner. The discriminator is updated with one gradient descent step, after which the generator is updated with two gradient descent steps. $\lambda$ is set as 5 for gradient penalty in $\mathcal{L}_{adv}^d$. $\alpha$, $\beta$ and $\gamma$ are set as 20, 100 and 20 respectively, which produces promising results in our experiment. The Adam solver is used for both the generator and discriminator with a batch size of 1.

**Evaluation Metrics.** Two metrics are used to quantify the difference between the predicted 3D facial voxel grid and the ground truth. 1) Mean Intersection-over-Union (IoU) [21]:

$$IoU = \frac{\sum_{i=1}^{N} [C(y_i > T) \times C(\hat{y}_i)]}{\sum_{i=1}^{N} [C(C(y_i > T) + C(\hat{y}_i))]} \qquad (12)$$

where $C(\cdot)$ is the indicator function, $y_i$ is the predicted occupancy state of the $i$th voxel, $\hat{y}_i$ is the corresponding ground truth, $T$ is the threshold for voxelization, and $N$ is the number of voxels in the grid. $T$ is set as 0.5 in our experiments. The higher the IoU value, the better the 3D facial geometry recovery.

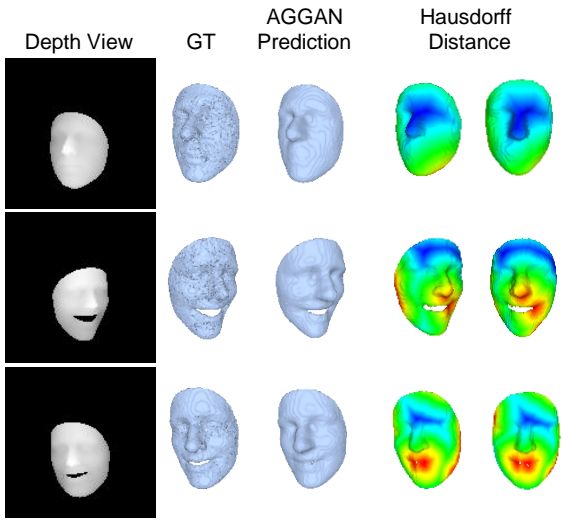

Fig. 5. Comparsion between the AGGAN prediction and the ground truth (GT). The Hausdorff distance between the GT and prediction is calculated and colorized on the predicted 3D face. Please note that the distance value increases from red to blue.

2) Mean value of standard Cross-Entropy loss (CE) [21]:

$$CE = -\frac{1}{N}\sum_{i=1}^{N}[\hat{y}_i\log(y_i) + (1-\hat{y}_i)\log(1-y_i)] \qquad (13)$$

where $N$, $y_i$ and $\hat{y}_i$ are the same as in (12). The lower the CE level is, the closer the 3D prediction to be either '0' or '1', which indicates a more robust and confident prediction.

### B. Results

IoU and CE values calculated on the predictions of LFPW, IBUG and AFW in 300W-LP are reported in Table I. Meanwhile, in Fig. 4 there are some visual results of the recovered 3D facial geometry for qualitative evaluation. As shown in Fig. 4, AGGAN can recover the 3D facial geometry well for different head poses, facial identities and expressions, and even when there are random noises or problematic holes in the given depth view. We also calculate and visualize the Hausdorff distance (two sets are close in the Hausdorff distance if every point of either set is close to some point of the other set) between the predicted voxel grid and its ground truth (see Fig. 5, the distance value increases from red to blue). To further prove the accuracy of the facial identity prediction, we show 3D results predicted from depth views with an identical facial identity but projected under different head poses in Fig. 6.

**Comparison with Existing Methods.** We compare AGGAN with some representative 3D surface reconstruction and non-rigid registration methods, including Marching Cubes (MC) [8], Screened Poisson Surface Reconstruction (SPSR) [11] and non-rigid ICP (NICP) [13]. For algorithms such as NICP that require connectivity, the Ball-Pivoting algorithm [1] is used to compute a triangle mesh interpolating the given facial point cloud. To get a promising result for NICP [13], we first applied ICP to rigidly align the facial template with the given facial point cloud, then initialized the non-rigid registration with hand-selected facial landmarks. Since each aforementioned method recons-tructs the 3D face in a distinct topology whose vertex amount and connectivity are different from each other, we cannot compare the methods using IoU and CE. Alternatively, we report the visual comparison results in Fig. 7. It can be seen 3D faces recovered by previous methods are severely distorted (Fig. 7), when the input depth view is in a large head pose, incomplete and with prominent artefacts. In contrast, AGGAN is much more roust to data imperfections in the depth view and able to generate a complete and smooth 3D facial geometry with facial identity and expression well preserved. What's more, as shown in the third and fifth row in Fig. 7, AGGAN can normally generate a 3D face smoother and denser than the ground truth since it predicts the probability of each voxel occupancy within the range of $[0, 1]$ continuously.

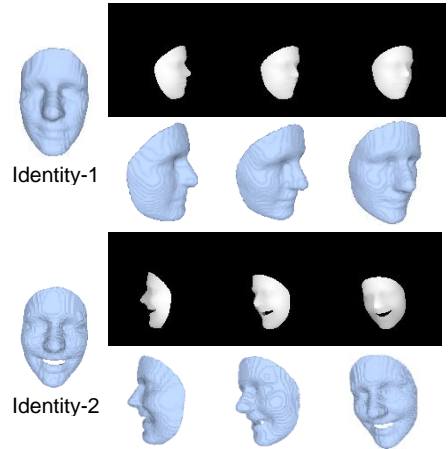

Fig. 6. Results of AGGAN predicted from depth views with an identical facial identity however with different head poses.

| Attention | Sparsity | IoU | CE |
|---|---|---|---|
| ✗ | ✗ | 0.9917 | 0.1151 |
| ✓ | ✗ | **0.9932** | **0.0940** |
| ✗ | ✓ | 0.9928 | 0.0995 |
| ✓ | ✓ | 0.9927 | 0.1004 |

**Ablation Analysis.** The importance of the sparsity constraint and the attention module is investigated. Specifically, four different AGGAN models which cover all possible combinations (with/without sparsity and with/without attention) of the two modules were trained on HELEN and tested on a subset of IBUG - a challenging dataset which contains facial images of very large head pose and facial expression. IoU and CE levels of these four models on the testing set are listed in Table II. It can be found that both sparsity and attention help improve AGGAN's prediction accuracy when they work independently. Moreover, the model with the attention module outputs the best result, which verifies the significance of attention in AGGAN and implies that there might conflict between sparsity and attention during the network learning process.

### C. Limitations and Prospect

A few artefacts can be observed around the facial boundary in the predictions of AGGAN. For example, as shown in Fig. 4, the inner mouth region cannot be fully recovered when the facial expression is a big open mouth. We think this is mainly due to that the voxel grid used is not dense enough. When voxel occupancy states were predicted mistakenly, the resulted due to that the voxel grid used is not dense enough. When voxel occupancy states were predicted mistakenly, the resulted artefacts would be obvious. This problem can be alleviated by using a denser voxel grid or applying a better prior to restrict the voxel occupancy state for forming a reasonable face, e.g. using a mean face with neutral facial expression as a template grid and driving AGGAN to predict the difference between the template and the target face. Although AGGAN has been validated on the synthetic data, it shows its potential for the application of real depth views captured from depth cameras. For example, as shown in the 3rd and 4th column of Fig. 4, AGGAN can recover 3D facial geometry accurately when there are random noises or even problematic holes (please note that these holes were not simulated in the training data) in the depth view. To fill in the gap between the synthetic data and real data, a promising direction is to train a network learning the common feature representation of the synthetic and real depth views. In this way, the synthetic data can be sufficiently utilized while much less real depth views will need to be collected

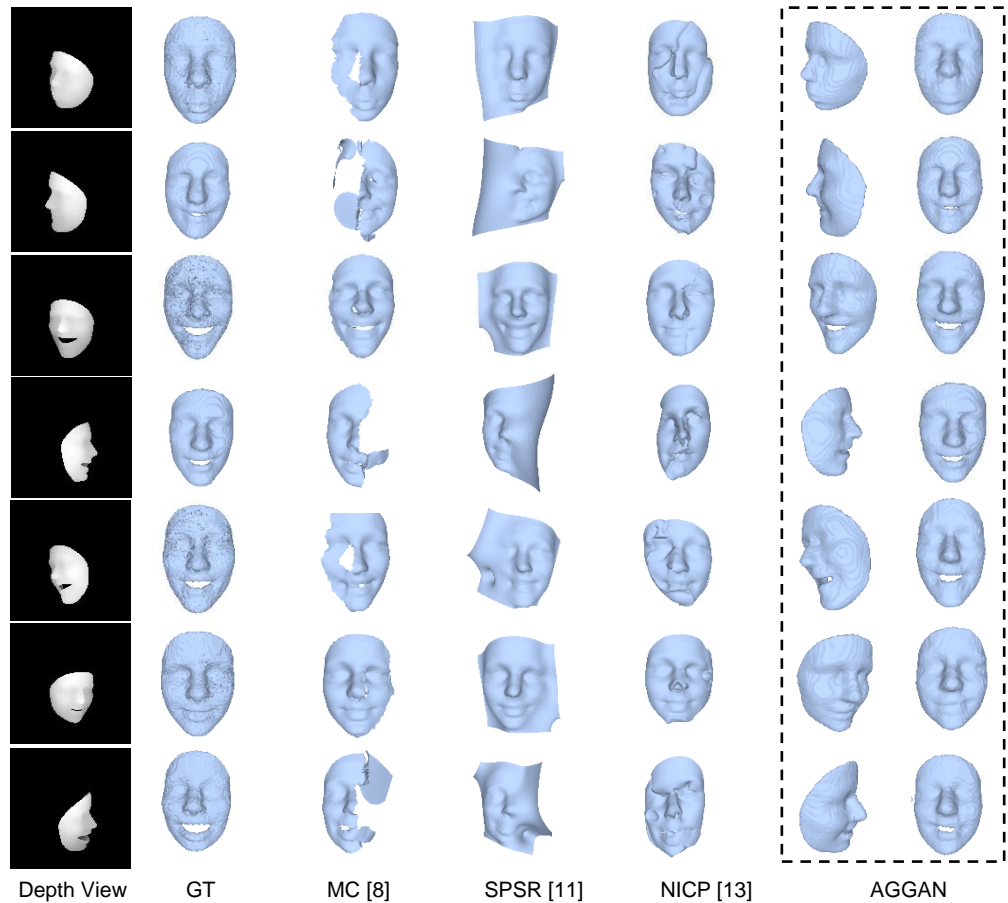

Fig. 7. Comparison between AGGAN and existing methods on challenging depth views.

## V. CONCLUSIONS

In this paper, we propose to model the ill-posed 2.5D facial depth-3D mapping with a novel attention-guided GAN structure – AGGAN in a data-driven manner. AGGAN is validated on synthetic depth views which cover a wide range of facial identities, expressions and head poses. When dealing with noisy and non-frontal facial depth views, AGGAN is still capable of recovering the 3D structure of the missing/occluded facial parts with facial identity and expression being accurately preserved, and thus significantly outperforms previous methods. Moreover, AGGAN is resilient to data imperfections in the depth view such as random noise and problematic holes, and hence has a potential of being applied to real depth views captured by depth cameras.